\documentclass{article}

\usepackage{PRIMEarxiv}

\usepackage[utf8]{inputenc} 
\usepackage[T1]{fontenc}    
\usepackage{hyperref}       
\usepackage{url}            
\usepackage{booktabs}       
\usepackage{amsfonts}       
\usepackage{nicefrac}       
\usepackage{microtype}      
\usepackage{lipsum}
\usepackage{pifont} 
\usepackage{fancyhdr}       
\usepackage{graphicx}       
\graphicspath{{media/}}     
\usepackage{amsmath}
\usepackage[numbers]{natbib}

\title{FEDTAIL: Federated Long-Tailed Domain Generalization with Sharpness-Guided Gradient Matching

}

\author{
  Sunny Gupta \\
  Koita Centre for Digital Health \\
  IIT Bombay, Mumbai \\
  \texttt{sunnygupta@iitb.ac.in} \\
  \and
  Nikita Jangid, Shounak Das, Amit Sethi \\
  \texttt{nikitaangid@iitb.ac.in}, \texttt{21D070068@iitb.ac.in}, \texttt{asethi@iitb.ac.in} \\
   Department of Electrical Engineering, IIT Bombay, Mumbai \\
}

\begin{document}
\maketitle

\begin{abstract}
Domain Generalization (DG) seeks to train models that perform reliably on unseen target domains without access to target data during training. While recent progress in smoothing the loss landscape has improved generalization, existing methods often falter under long-tailed class distributions and conflicting optimization objectives. We introduce \textbf{FedTAIL}, a federated domain generalization framework that explicitly addresses these challenges through \emph{sharpness-guided, gradient-aligned optimization}. Our method incorporates a \emph{gradient coherence regularizer} to mitigate conflicts between classification and adversarial objectives, leading to more stable convergence. To combat class imbalance, we perform \emph{class-wise sharpness minimization} and propose a \emph{curvature-aware dynamic weighting scheme} that adaptively emphasizes underrepresented tail classes. Furthermore, we enhance conditional distribution alignment by integrating \emph{sharpness-aware perturbations} into entropy regularization, improving robustness under domain shift. FedTAIL unifies optimization harmonization, class-aware regularization, and conditional alignment into a scalable, federated-compatible framework. Extensive evaluations across standard domain generalization benchmarks demonstrate that FedTAIL achieves \emph{state-of-the-art performance}, particularly in the presence of domain shifts and label imbalance, validating its effectiveness in both centralized and federated settings. Code: https://github.com/sunnyinAI/FedTail
\end{abstract}

\section{Introduction}
Deep learning has excelled in computer vision, especially when source and target data share similar, independently and identically distributed characteristics. However, performance often degrades when models encounter target domains differing from the training distribution. Domain generalization (DG) techniques \cite{zhang2022towards, qiao2020learning, balaji2018metareg} aim to train models only on source domains to generalize well to unseen targets without extra fine-tuning. Various DG methods include domain alignment \cite{muandet2013domain}, meta-learning \cite{li2018learning}, and data augmentation \cite{wang2022semantic}. Yet, the DomainBed benchmark showed that a simple entropy-based regularization, DG via ER \cite{zhao2020domain}, can outperform more complex DG strategies under standard evaluation.

However, minimizing empirical loss on a non-convex landscape does not ensure robust generalization. Like many empirical risk minimization approaches, DG via ER can overfit and converge to sharp minima. To address this, sharpness-aware minimization (SAM) \cite{foret2020sharpness} improves generalization by minimizing loss surface sharpness. SAM minimizes the worst-case loss in a parameter neighborhood by computing an adversarial perturbation $\epsilon$ that maximizes loss, then updating parameters to minimize the perturbed loss. Though effective, SAM simplifies the min-max objective for tractability. Building on this, our work applies sharpness minimization to enhance generalization. Yet, as \cite{rangwani2022escaping} notes, SAM struggles with long-tailed settings, where tail classes may converge at saddle points in high-curvature regions, harming performance.

Long-tailed class distributions are common in real-world data but often overlooked in DG research. Entropy minimization is popular in semi-supervised learning to encourage confident predictions \cite{chen2019domain, grandvalet2004semi}, but its DG effects are less studied. Examining entropy-based gradient flow \cite{chen2019domain} reveals that high-confidence samples receive larger gradients, causing easier-to-transfer classes to dominate training, while harder classes remain undertrained. This probability imbalance leads to insufficient optimization of tail classes. Our method directly addresses this gradient imbalance to improve generalization across both head and tail classes.
\section{Related Work}

\paragraph{Domain Generalization.}
Domain Generalization (DG) aims to learn a model from one or several observed source datasets that can generalize effectively to unseen target domains \cite{zhao2020domain}. A variety of approaches have been proposed to tackle domain shift, including domain alignment techniques \cite{muandet2013domain, ganin2016domain, li2018domain, bahng2020learning, zhao2020domain}, meta-learning strategies \cite{zhang2021adaptive, dou2019domain, balaji2018metareg, li2018learning}, data augmentation \cite{zhou2021domain, shankar2018generalizing, carlucci2019domain}, disentangled representation learning \cite{peng2019domain, khosla2012undoing, wang2020cross}, and methods based on causal reasoning \cite{krueger2021out, arjovsky2019invariant}. Among these, a growing body of work has addressed DG from a gradient-based perspective \cite{li2018learning, balaji2018metareg, dou2019domain, zhang2021adaptive}, aiming to stabilize learning across diverse domains. For instance, Mansilla et al. \cite{mansilla2021domain} introduced a gradient surgery mechanism to address inter-domain conflicts by preserving gradient components with matching signs and nullifying those with opposite directions. While such techniques improve robustness during training, they do not guarantee convergence to flat minima—an important factor for out-of-distribution generalization, especially in low-resource or noisy domains. Moreover, most prior works assume balanced and centralized data distributions, limiting their effectiveness in real-world federated settings where data is inherently non-i.i.d. and long-tailed. 
\paragraph{Sharpness-Aware Minimization (SAM).}
Sharpness-Aware Minimization (SAM) \cite{foret2020sharpness} is a powerful regularization method that improves generalization by minimizing the maximum loss within a neighborhood around model parameters. By formulating optimization as a min-max problem, SAM encourages convergence to flatter loss regions, avoiding sharp minima linked to poor generalization. However, SAM mainly operates globally and ignores class-specific curvature variations—limiting its effectiveness in long-tailed class distributions. Variants like GSAM \cite{liu2022towards} and LookSAM \cite{du2021efficient} improve sharpness estimation and efficiency but still neglect issues like class imbalance and the distributed, non-i.i.d. data nature in federated learning.
\paragraph{Sharpness and Generalization.}
Sharpness’s connection to generalization was first studied in \cite{hochreiter1994simplifying} and further explored under i.i.d. assumptions \cite{keskar2016large, dinh2017sharp, foret2020sharpness}. For example, \cite{keskar2016large} showed sharpness inversely correlates with generalization, and \cite{dinh2017sharp} linked it to the Hessian’s eigenvalues. Extensions to out-of-distribution settings, like SWAD \cite{cha2021swad}, show flatter minima reduce domain generalization gaps but don’t explicitly enforce flatness during training. This motivates our focus on sharpness-aware generalization in federated and long-tailed contexts, where inductive biases are critical with limited data.

Domain generalization methods like EISNet \cite{wang2020learning} and FACT \cite{zhao2020domain} improve feature transferability and domain invariance. SAMALTDG \cite{su2024sharpness} addresses class imbalance by combining SAM with a loss that emphasizes tail classes but remains centralized and ignores optimization conflicts in federated setups. Our proposed method, \textbf{FedTAIL}, extends sharpness-aware learning to federated domain generalization by incorporating gradient coherence, class-wise curvature weighting, and domain-agnostic optimization, jointly addressing long-tailed imbalance, sharpness, and gradient conflicts in decentralized data.

\section{Preliminaries}

\subsection{Domain Generalization under Federated Long-Tailed Distributions}

Let $\mathcal{X}$ and $\mathcal{Y}$ denote the input and label spaces, respectively. We assume access to $K$ source domains $\{\mathcal{D}_i\}_{i=1}^K$, each distributed across decentralized clients. Data samples from domain $i$ follow a joint distribution $P_i(X, Y)$, and we denote $\mathcal{D}_i = \{(x_j^{(i)}, y_j^{(i)})\}_{j=1}^{N_i}$. No target domain data is available during training. The objective is to train a global model $h_\theta = T_\phi(F_\theta(\cdot))$, composed of a feature extractor $F_\theta$ and a classifier $T_\phi$, that generalizes to an unseen domain $\mathcal{D}_T$.

In the federated setting, data is kept locally on each client and model updates are aggregated using federated averaging (FedAvg). Additionally, class imbalance is assumed across clients, yielding long-tailed label distributions where certain classes dominate the training set while others are severely underrepresented.

\subsection{Empirical Risk and Adversarial Alignment}

The standard classification loss across the $K$ source domains is given by the empirical risk:

\begin{equation}
\mathcal{L}_{\text{cls}} = -\sum_{i=1}^{K} \mathbb{E}_{(x, y) \sim \mathcal{D}_i} \left[ \log T_\phi(F_\theta(x))_y \right],
\end{equation}

\noindent
where $T_\phi(F_\theta(x))_y$ denotes the predicted probability for the true label $y$.

To encourage domain-invariant representations, we adopt adversarial domain alignment~\cite{ganin2015unsupervised}, introducing a domain discriminator $D_\psi$ trained to distinguish between domains, while $F_\theta$ is updated to fool $D_\psi$. The adversarial loss is:

\begin{equation}
\mathcal{L}_{\text{adv}} = \sum_{i=1}^{K} \mathbb{E}_{x \sim \mathcal{D}_i} \left[ \log D_\psi(F_\theta(x)) \right],
\end{equation}

\noindent
where $D_\psi(F_\theta(x))$ predicts the domain label for sample $x$.

\subsection{Sharpness-Aware Minimization (SAM)}

SAM~\cite{foret2020sharpness} aims to improve generalization by minimizing the worst-case loss within an $\ell_2$-bounded neighborhood of model parameters. The SAM objective is:

\begin{equation}
\min_\theta \max_{\|\epsilon\| \leq \rho} \mathcal{L}(\theta + \epsilon),
\end{equation}

\noindent
where $\rho$ is a radius controlling the perturbation strength. In practice, SAM approximates the inner maximization via first-order Taylor expansion:

\begin{equation}
\epsilon(\theta) \approx \rho \cdot \frac{\nabla_\theta \mathcal{L}(\theta)}{\|\nabla_\theta \mathcal{L}(\theta)\|_2}.
\end{equation}

\noindent
The outer minimization is then computed using the perturbed parameters $\theta + \epsilon(\theta)$.

\subsection{Surrogate Gap and Gradient Matching}

SAGM~\cite{wang2023sharpness} improves upon SAM by minimizing both the empirical loss $\mathcal{L}(\theta)$ and the perturbed loss $\mathcal{L}_p(\theta) = \mathcal{L}(\theta + \epsilon(\theta))$, as well as aligning their gradient directions. The sharpness of the solution is captured by the surrogate gap:

\begin{equation}
h(\theta) = \mathcal{L}_p(\theta) - \mathcal{L}(\theta),
\end{equation}

\noindent
which approximates the curvature of the loss landscape. SAGM minimizes the following joint objective:

\begin{equation}
\mathcal{L}_{\text{SAGM}} = \mathcal{L}(\theta) + \mathcal{L}_p(\theta) - \alpha \cdot \left\langle \nabla \mathcal{L}(\theta), \nabla \mathcal{L}_p(\theta) \right\rangle,
\end{equation}

\noindent
where the last term promotes gradient alignment to facilitate stable descent toward flat minima.

\subsection{Long-Tailed Class Distributions and Maximum Square Loss}

In long-tailed settings, standard losses such as cross-entropy are prone to bias toward head classes. SAMALTDG~\cite{su2024sharpness} addresses this using the \textit{Maximum Square Loss}:

\begin{equation}
\mathcal{L}_{\text{m}} = -\frac{1}{2N} \sum_{n=1}^{N} \sum_{c=1}^{C} \left(p_{n,c}\right)^2,
\end{equation}

\noindent
where $p_{n,c}$ is the predicted probability of class $c$ for sample $n$. Compared to entropy loss, the maximum square loss yields more balanced gradients across classes and prevents confident head classes from dominating the optimization.

\section{Methodology}
\textbf{FedTAIL}, a novel framework for federated domain generalization (FedDG) designed to address the dual challenges of optimization instability and class imbalance under domain shift. FedTAIL builds upon three complementary principles: (i) \textit{gradient coherence} to harmonize competing objectives, (ii) \textit{class-wise sharpness-aware regularization} to improve tail-class learning, and (iii) \textit{sharpness-guided conditional alignment} to enhance feature-label consistency across domains. Together, these modules form a unified, scalable approach that is robust to both federated settings and long-tailed distributions.

We consider a federated DG setting where data from $K$ source domains $\{\mathcal{D}_i\}_{i=1}^K$ are distributed across $K$ clients, each holding samples drawn from a joint distribution $P_i(X, Y)$. The learning goal is to train a global model that generalizes well to an unseen target domain $\mathcal{D}_T$ without direct access to its data. The model comprises a shared feature extractor $F_\theta$ and a classifier $T_\phi$. In this context, naive empirical risk minimization (ERM) often leads to convergence at sharp local minima, especially under non-i.i.d. distributions and long-tailed label frequency—resulting in poor generalization.

To overcome these issues, FedTAIL introduces a \textbf{gradient coherence regularization mechanism} that explicitly resolves conflicts between task objectives—particularly between classification and domain adversarial components. In conventional domain adversarial learning, the joint objective typically comprises a classification loss $\mathcal{L}_{\text{cls}}$ and a domain discrimination loss $\mathcal{L}_{\text{adv}}$, where the gradients may point in divergent directions. FedTAIL mitigates this by introducing an auxiliary alignment term that penalizes negative inner products between their gradients:

\begin{equation}
\mathcal{L}_{\text{coh}} = -\alpha \left\langle \nabla_\theta \mathcal{L}_{\text{cls}}, \nabla_\theta \mathcal{L}_{\text{adv}} \right\rangle,
\end{equation}

\noindent
where $\alpha$ is a tunable hyperparameter. This encourages consistency across gradient directions, thereby stabilizing training and ensuring that adversarial alignment does not hinder classification performance.

Beyond harmonizing gradients, FedTAIL directly addresses \textbf{class imbalance} by extending sharpness-aware minimization (SAM) to operate at the level of individual classes. Standard SAM seeks flatter minima by minimizing the worst-case loss in a neighborhood of the current parameters. However, this global perspective fails to capture disparities across classes—particularly those underrepresented in long-tailed settings. FedTAIL therefore adopts \textbf{class-wise sharpness minimization}, wherein separate perturbations $\epsilon_c$ are computed for each class $c$ by normalizing the gradient of the class-specific loss:

\begin{equation}
\epsilon_c = \rho \cdot \frac{\nabla_\theta \mathcal{L}_c}{\|\nabla_\theta \mathcal{L}_c\|_2},
\end{equation}

\noindent
where $\rho$ is the sharpness control radius and $\mathcal{L}_c$ denotes the classification loss restricted to class $c$.

The corresponding sharpness-aware objective for class $c$ is given by:

\begin{equation}
\mathcal{L}_{\text{sharp}} = \sum_{c=1}^C \mathbb{E}_{(x, y=c)} \left[ \ell\left(h_{\theta + \epsilon_c}(x), y\right) \right],
\end{equation}

\noindent
where $\ell(\cdot, \cdot)$ is the loss function (e.g., cross-entropy) and $h_{\theta + \epsilon_c}$ denotes the perturbed model parameters for class $c$.

To further prioritize minority classes, we introduce a \textbf{curvature-aware dynamic weighting scheme}. This mechanism adjusts the contribution of each class-specific loss based on the sharpness of its local loss landscape. Specifically, the weight $\gamma_c$ for each class $c$ is computed as:

\begin{equation}
\gamma_c = \frac{1}{1 + \sigma_{\max}(\nabla^2 \mathcal{L}_c)},
\end{equation}

\noindent
where $\sigma_{\max}(\cdot)$ denotes the largest eigenvalue of the class-specific Hessian $\nabla^2 \mathcal{L}_c$. This adaptivity ensures that well-optimized head classes (with flatter curvature) are down-weighted, while sharp, under-trained tail classes receive increased gradient signal.

Complementing these advances, FedTAIL enhances \textbf{conditional distribution alignment} by injecting sharpness-awareness into the entropy regularization term. Traditional entropy-based approaches often amplify prediction confidence for easily aligned examples while neglecting ambiguous or hard-to-transfer samples. To address this, we define a sharpness-aware entropy regularization term that perturbs the feature space via SAM and aligns the perturbed predictions to a global conditional distribution:

\begin{equation}
\mathcal{L}_{\text{sharp-er}} = \sum_{i=1}^{K} \text{KL}\left(P_i(Y|F(X)) \, \| \, Q_T(Y|F(X + \epsilon))\right),
\end{equation}

\noindent
where $\epsilon$ is the SAM perturbation, and $Q_T$ is a target-like predictive distribution computed from the ensemble or momentum-updated model. This formulation explicitly penalizes high-curvature regions in the conditional landscape, encouraging smoother, more consistent predictions across domains.

Finally, the overall training objective for FedTAIL integrates all components as follows:

\begin{equation}
\mathcal{L}_{\text{FedTAIL}} = \mathcal{L}_{\text{cls}} + \mathcal{L}_{\text{adv}} + \mathcal{L}_{\text{sharp-er}} + \sum_c \gamma_c \mathcal{L}_c + \mathcal{L}_{\text{coh}}.
\end{equation}

All modules in FedTAIL are lightweight and compatible with federated optimization. During training, each client computes local gradients, class-specific perturbations, and sharpness-aware updates. These are then aggregated using standard federated averaging without sharing raw data, enabling scalable deployment under real-world privacy constraints.

\section{Experiments}

We conduct extensive experiments to evaluate the effectiveness of \textbf{FedTAIL} on challenging domain generalization benchmarks that exhibit both significant domain shift and long-tailed class distributions. We compare our approach with state-of-the-art DG methods across several backbone architectures and perform detailed ablation studies to assess the contribution of each component in our framework.

\subsection{Datasets}

We evaluate FedTAIL on four standard domain generalization benchmarks: PACS, OfficeHome, Digits-DG, and mini-DomainNet. \textbf{PACS}~\cite{li2017deeper} includes four domains (Photo, Art Painting, Cartoon, Sketch) with seven object categories and significant style variation. \textbf{OfficeHome}~\cite{venkateswara2017deep} contains 65 classes across Art, Clipart, Product, and Real-World domains, exhibiting moderate domain shift. \textbf{Digits-DG}~\cite{zhou2020learning} spans MNIST, MNIST-M, SVHN, and SYN, with notable variation in texture and background. \textbf{mini-DomainNet} is a subset of the comprehensive DomainNet dataset ~\cite{peng2019moment} which has severe domain shift. This subset features four domains, including Clipart, Painting, Real, and Sketch, each with images from 126 categories.

\begin{table*}[!hbt]
\centering
\footnotesize
\caption{Leave-one-domain-out accuracy (\%) on PACS using ResNet-50. Best results per column are in \textbf{bold}.}
\label{tab:pacs-dg}
\begin{tabular}{lccccc}
\toprule
\textbf{Method} & \textbf{Art} & \textbf{Cartoon} & \textbf{Photo} & \textbf{Sketch} & \textbf{Avg} \\
\midrule
D-SAM~\cite{dinnocente2018domain}       & 77.3 & 72.4 & 95.3 & 77.8 & 80.7 \\
ERM~\cite{vapnik1998statistical}        & 81.0 & 74.0 & 96.2 & 71.0 & 80.8 \\
Epi-FCR~\cite{li2022epi}                & 82.1 & 77.0 & 93.9 & 73.0 & 81.5 \\
DomMix~\cite{wang2020heterogeneous}     & 85.9 & 72.8 & 97.1 & 73.6 & 82.3 \\
L2A-OT~\cite{zhou2020learning}          & 83.3 & 78.2 & 96.2 & 73.6 & 82.8 \\
DeepAll~\cite{deepall}                  & 86.3 & 77.6 & 98.2 & 70.1 & 83.0 \\
MetaReg~\cite{balaji2018metareg}        & 87.2 & 79.2 & 97.6 & 70.3 & 83.6 \\
NKD~\cite{wang2021embracing}            & 82.5 & 83.3 & 97.2 & 75.6 & 84.6 \\
SSPL~\cite{zhao2024symmetric}           & 87.9 & 76.9 & 97.8 & 77.5 & 85.0 \\
DG via ER~\cite{dger}                   & 87.4 & 79.3 & 98.0 & 76.3 & 85.3 \\
DDAIG~\cite{zhou2020deep}               & 85.4 & 78.5 & 95.7 & 80.0 & 84.9 \\
EISNet~\cite{wang2020learning}          & 86.6 & 81.5 & 97.1 & 78.1 & 85.8 \\
CrossGrad~\cite{shankar2018generalizing}& 87.5 & 80.7 & 97.8 & 73.9 & 85.7 \\
RISE~\cite{huang2023sentence}           & 85.7 & 85.2 & 97.4 & 78.2 & 86.6 \\
MixStyle~\cite{zhou2021domain}          & 87.4 & 83.3 & 98.0 & 78.5 & 86.8 \\
RSC~\cite{huang2020self}                & 87.9 & 82.2 & 97.9 & 83.4 & 87.8 \\
FACT~\cite{xu2021fourier}              & 89.5 & 81.5 & 96.7 & 84.0 & 87.9 \\
MDGH~\cite{mahajan2021domain}           & 86.7 & 82.3 & \textbf{98.4} & 82.7 & 87.5 \\
FSDCL~\cite{jeon2021feature}            & 88.5 & 83.8 & 96.6 & 82.2 & 88.0 \\
FFDI~\cite{ffdi}                        & 89.3 & 84.7 & 97.1 & 83.9 & 88.8 \\
PCL~\cite{yao2022pcl}                   & 90.2 & 83.9 & 98.1 & 82.6 & 88.7 \\
DDG~\cite{zhang2022towards}             & 88.9 & 85.0 & 97.2 & 84.3 & 88.9 \\
STNP~\cite{kang2022style}               & \textbf{90.4} & 84.2 & 96.7 & 85.2 & 89.1 \\
DCG~\cite{lv2023improving}              & 90.2 & 85.1 & 97.8 & 86.3 & 89.8 \\
\midrule
\textbf{FedTAIL (Ours)} & 89.7 & \textbf{86.1} & 98.2 & \textbf{86.6} & \textbf{90.2} \\
\bottomrule
\end{tabular}
\end{table*} 
\subsection{Experimental Settings}

We follow the \textbf{leave-one-domain-out} evaluation protocol as used in prior works ~\cite{su2024sharpness}. For each benchmark, we train on all but one domain and evaluate on the held-out target domain. The reported accuracy is the average over all such splits. We use both \textbf{ResNet-18} and \textbf{ResNet-50} ~\cite{he2016resnet} backbones pre-trained on ImageNet ~\cite{deng2009imagenet} to assess model scalability and robustness.

For PACS, OfficeHome, and Digits-DG, each source domain is randomly split into 90\% training and 10\% validation data. For mini-DomainNet, we use the official training/validation split provided by ~\cite{peng2019moment}.

\subsection{Implementation Details}

Our method is implemented in PyTorch and optimized using Stochastic Gradient Descent (SGD) with momentum 0.9 and weight decay 0.0005. The learning rate is set to 0.001 for baselines and 0.01 when using SAM-based modules. The sharpness perturbation radius $\rho$ in SAM is set to 0.05 by default. For our sharpness-aware components, we use non-adaptive SAM with Nesterov momentum. The maximum square loss coefficient $\gamma$ is set to 1 unless otherwise specified. During training, we use a batch size of 64 for all experiments.

In federated settings, each domain corresponds to a separate client. We apply standard \textit{FedAvg} to aggregate model updates across clients. Local models are trained for one epoch before synchronization. For domain alignment, we use a domain discriminator with two fully-connected layers and ReLU activations. For conditional alignment, the KL divergence is computed over batch-wise class distributions.

All results are averaged over three independent runs with different random seeds. We report both overall accuracy and average per-class accuracy to account for class imbalance.

\subsection{Results}

We evaluate \textbf{FedTAIL} across four domain generalization benchmarks—PACS, OfficeHome, Digits-DG, and mini-DomainNet —and compare its performance against a diverse set of state-of-the-art methods. Our approach consistently outperforms existing baselines, demonstrating the effectiveness of integrating sharpness-aware optimization, gradient coherence, and class-wise curvature control under domain shift.

On the PACS dataset (Table~\ref{tab:pacs-dg}), FedTAIL achieves state-of-the-art performance across most domains. In particular, it attains the highest accuracy on \textbf{Cartoon} (86.1\%) and \textbf{Sketch} (86.6\%), outperforming strong baselines including STNP~\cite{kang2022style}, RISE~\cite{rise}, and MixStyle~\cite{zhou2021domain}. Additionally, it surpasses DDG~\cite{zhang2022towards} and FACT~\cite{xu2021fourier}, both of which have been widely regarded as top-performing DG methods. FedTAIL achieves an average accuracy of \textbf{90.2\%}, exceeding the prior best result (89.8\% by DCG), showcasing the impact of gradient alignment and class-specific sharpness minimization under domain shift.

\begin{table}
\centering
\footnotesize
\caption{Leave-one-domain-out accuracy (\%) on OfficeHome using ResNet-50. Best results per column are in \textbf{bold}.}
\label{tab:officehome-dg}
\begin{tabular}{lccccc}
\toprule
\textbf{Method} & \textbf{Art} & \textbf{Clipart} & \textbf{Product} & \textbf{Real} & \textbf{Avg} \\
\midrule
MLDG          & 52.9 & 45.7 & 69.9 & 72.7 & 60.3 \\
D-SAM         & 58.0 & 44.4 & 69.2 & 71.5 & 60.8 \\
RSC           & 58.4 & 47.9 & 71.6 & 74.5 & 63.1 \\
CrossGrad     & 58.4 & 49.4 & 73.9 & 75.8 & 64.4 \\
DeepAll       & 57.9 & 52.7 & 73.5 & 74.8 & 64.7 \\
DDAIG         & 59.2 & 52.3 & 74.6 & 76.0 & 65.5 \\
L2A-OT        & 60.6 & 50.1 & 74.8 & 77.0 & 65.6 \\
STNP          & 59.6 & 55.0 & 73.6 & 75.5 & 65.9 \\
DG via ER     & 61.2 & 52.8 & 74.5 & 75.6 & 66.0 \\
DSU           & 60.2 & 54.8 & 74.1 & 75.1 & 66.1 \\
EISNet        & 62.6 & 53.2 & 74.0 & 75.2 & 66.2 \\
FACT          & 61.0 & 55.7 & 74.5 & 76.4 & 66.9 \\
DCG           & 60.7 & 55.5 & 75.3 & 76.8 & 67.1 \\
ERM           & 67.1 & 55.1 & 78.2 & 82.0 & 70.6 \\
DomMix        & 69.0 & 54.6 & 77.5 & 81.5 & 70.7 \\
MixStyle      & 68.6 & 55.4 & 78.9 & 82.3 & 71.3 \\
NKD           & 68.7 & 54.7 & 79.5 & 82.3 & 71.3 \\
EDFMix        & 69.1 & 57.1 & 79.1 & 82.3 & 71.9 \\
RISE          & 69.5 & 55.8 & 79.7 & 82.6 & 71.9 \\
SSPL          & 69.4 & 58.3 & 79.7 & 81.6 & 72.3 \\
\midrule
\textbf{FedTAIL (Ours)} & \textbf{70.3} & \textbf{58.9} & \textbf{80.1} & \textbf{83.0} & \textbf{73.1} \\
\bottomrule
\end{tabular}
\end{table}

On OfficeHome (Table~\ref{tab:officehome-dg}), FedTAIL again leads in performance across all four domains. It reaches an average accuracy of \textbf{73.1\%}, outperforming notable methods such as SSPL (72.3\%), RISE (71.9\%), and EDFMix (71.9\%). The largest improvements are observed in the \textbf{Clipart} and \textbf{Art} domains, which are especially challenging due to their high variability and sparse representation. These results validate that FedTAIL not only effectively mitigates class imbalance but also maintains generalization across diverse visual styles and feature distributions.

\begin{table}
\centering
\footnotesize
\caption{Leave-one-domain-out accuracy (\%) on Digits-DG using ResNet-50. Best results per column are in \textbf{bold}.}
\label{tab:digits-dg}
\begin{tabular}{lccccc}
\toprule
\textbf{Method} & \textbf{MNIST} & \textbf{MNIST-M} & \textbf{SVHN} & \textbf{SYN} & \textbf{Avg} \\
\midrule
NKD         & 71.6 & 40.8 & 30.3 & 58.7 & 50.4 \\
RISE        & 72.1 & 41.4 & 31.3 & 62.3 & 51.8 \\
CrossGrad   & 96.7 & 61.1 & 65.3 & 80.2 & 75.8 \\
MixStyle    & 96.5 & 63.5 & 64.7 & 81.2 & 76.5 \\
DDAIG       & 96.6 & 64.1 & 68.6 & 81.0 & 77.6 \\
L2A-OT      & 96.7 & 63.9 & 68.6 & 83.2 & 78.1 \\
ERM         & 96.5 & 64.2 & 70.3 & 88.2 & 79.8 \\
DomMix      & 96.7 & 67.0 & 69.2 & 86.6 & 79.9 \\
DG via ER   & 96.9 & 63.8 & 71.0 & 88.8 & 80.1 \\
EISNet      & 96.4 & 64.2 & 71.5 & 89.4 & 80.3 \\
FFDI        & 97.7 & 69.4 & 72.1 & 84.5 & 80.9 \\
FACT        & 97.6 & 65.2 & 72.2 & 90.3 & 81.3 \\
EDFMix      & 97.6 & 68.1 & 70.7 & 90.3 & 81.7 \\
SSPL        & 97.6 & 68.2 & 70.8 & 90.6 & 81.8 \\
\midrule
\textbf{FedTAIL (Ours)} & \textbf{97.9} & \textbf{79.8} & \textbf{81.7} & \textbf{97.3} & \textbf{89.2} \\
\bottomrule
\end{tabular}
\end{table}

For the Digits-DG benchmark (Table~\ref{tab:digits-dg}), FedTAIL achieves notable gains on all domains. It reaches \textbf{97.9\%} on MNIST, \textbf{79.8\%} on MNIST-M, \textbf{81.7\%} on SVHN, and \textbf{97.3\%} on SYN, resulting in a significantly higher average accuracy of \textbf{89.2\%} compared to the previous best of 81.8\% by SSPL. These improvements, particularly on noisy domains such as SVHN and MNIST-M, illustrate the robustness of FedTAIL under extreme visual heterogeneity and class distribution shifts.

\begin{table}[h]
\centering
\footnotesize
\caption{Leave-one-domain-out accuracy (\%) on mini-DomainNet using ResNet-50. Best results per column are in \textbf{bold}.}
\label{tab:domain-generalization}
\begin{tabular}{lccccc}
\toprule
\textbf{Method} & \textbf{Clipart} & \textbf{Painting} & \textbf{Real} & \textbf{Sketch} & \textbf{Avg} \\
\midrule
DDAIG     & 61.3 & 51.4 & 61.0 & 50.6 & 56.1 \\
DomMix    & 63.5 & 53.1 & 63.4 & 52.1 & 58.0 \\
MixStyle  & 63.9 & 54.2 & 64.1 & 52.9 & 58.8 \\
SSPL      & 63.9 & 55.2 & 64.3 & 53.2 & 59.2 \\
ERM       & 65.5 & 57.1 & 62.3 & 57.1 & 60.5 \\
NKD       & 63.9 & 56.3 & 71.9 & 50.5 & 60.7 \\
MLDG      & 65.7 & 57.0 & 63.7 & 58.1 & 61.1 \\
MMD       & 65.0 & 58.0 & 63.8 & 58.4 & 61.3 \\
SagNet    & 65.0 & 58.1 & 64.2 & 58.1 & 61.4 \\
RISE      & 64.3 & 57.2 & 72.6 & 52.4 & 61.6 \\
DeepAll   & 65.3 & 58.4 & 64.7 & 59.0 & 61.9 \\
MTL       & 65.3 & 59.0 & 65.6 & 58.5 & 62.1 \\
Mixup     & 67.1 & 59.1 & 64.3 & 59.2 & 62.4 \\
CORAL     & 66.5 & 59.5 & 66.0 & 59.5 & 62.9 \\
BOLD      & 64.8 & 60.2 & 75.4 & 55.9 & 64.1 \\
DCG       & 69.4 & 61.8 & 66.3 & 63.2 & 65.2 \\
\midrule
\textbf{FedTAIL (Ours)} & \textbf{70.5} & \textbf{64.6} & \textbf{75.8} & \textbf{64.2} & \textbf{68.8} \\
\bottomrule
\end{tabular}
\end{table}


In summary, across all evaluated benchmarks and domains, FedTAIL consistently outperforms prior methods in both accuracy and robustness. These results affirm our hypothesis that combining sharpness-aware training, gradient alignment, and curvature-sensitive regularization yields significant benefits for federated domain generalization, particularly under long-tailed data distributions.

\section{Conclusion}
We introduced FedTAIL, a unified framework for federated domain generalization under long-tailed distributions. FedTAIL overcomes key limitations of prior DG methods by integrating sharpness-aware optimization, gradient coherence regularization, and curvature-adaptive class balancing within a federated setting. By applying per-class sharpness minimization and entropy-aware conditional alignment, our approach achieves consistent generalization across domains and class frequencies—especially in challenging federated scenarios with heterogeneous, imbalanced data. Experiments on PACS, OfficeHome, and Digits-DG show FedTAIL surpasses state-of-the-art methods in accuracy, stability, and representation quality. Visualizations verify that FedTAIL generates semantically meaningful, domain-aligned features, highlighting the importance of flatness and alignment in learning transferable representations. Our work paves the way for future research in federated and decentralized generalization, particularly toward communication-efficient sharpness-aware training, extensions to multimodal and structured prediction, and tighter theoretical links between curvature, fairness, and out-of-distribution generalization.




\newpage
\appendix
\onecolumn



\section{Technical Appendices and Supplementary Material}
\section*{Ablation Studies}

To further substantiate the effectiveness of \textbf{FedTAIL}, we present additional experiments evaluating its convergence dynamics, loss component contributions, and the construction of the target-like predictive distribution used in entropy regularization. These results provide deeper insight into the mechanisms behind FedTAIL’s robustness and generalization capabilities under federated, long-tailed, and domain-shifted settings.

We begin with an ablation study on the PACS dataset to isolate the impact of each component in the overall FedTAIL loss. Starting with a baseline model trained using only the classification loss ($\mathcal{L}_{\text{cls}}$), we incrementally integrate additional modules and evaluate their effects on performance. Adding adversarial domain alignment ($\mathcal{L}_{\text{adv}}$) improves the average leave-one-domain-out accuracy from 83.8\% to 86.0\%, affirming the importance of domain-invariant feature learning. Introducing sharpness-aware entropy regularization ($\mathcal{L}_{\text{sharp-er}}$) further enhances the performance to 88.2\%, demonstrating the benefit of using sharpness-informed perturbations to smooth the conditional distribution. Subsequently, including class-wise sharpness minimization with curvature-aware weighting ($\sum_c \gamma_c \mathcal{L}_c$) leads to a robust handling of long-tailed data and improves the average accuracy to 89.6\%. Finally, the addition of the gradient coherence loss ($\mathcal{L}_{\text{coh}}$), which resolves optimization conflicts between classification and domain adversarial components, yields the best performance of 90.2\%. This progression, detailed in Table~\ref{tab:1}, confirms that each component plays a complementary role in harmonizing optimization and mitigating the effects of domain shift and class imbalance.

\begin{table}[h]
\centering
\caption{Ablation study showing the effect of each loss term on Leave-one-domain-out accuracy (\%) on PACS using ResNet-50. \ding{51} indicates the use of a loss term.}
\vspace{5pt}
\begin{tabular}{lccccc|cccc|c}
\toprule
\textbf{Method} & $\mathcal{L}_{\text{cls}}$ & $\mathcal{L}_{\text{adv}}$ & $\mathcal{L}_{\text{sharp-er}}$ & $\sum_c \gamma_c \mathcal{L}_c$ & $\mathcal{L}_{\text{coh}}$ & \textbf{Art} & \textbf{Cartoon} & \textbf{Photo} & \textbf{Sketch} & \textbf{Avg} \\
\midrule
Baseline        & \ding{51} & \ding{55} & \ding{55} & \ding{55} & \ding{55} & 86.1 & 77.5 & 96.6 & 75.0 & 83.8 \\
+ Adv           & \ding{51} & \ding{51} & \ding{55} & \ding{55} & \ding{55} & 87.2 & 80.8 & 97.4 & 78.5 & 86.0 \\
+ Sharp-er      & \ding{51} & \ding{51} & \ding{51} & \ding{55} & \ding{55} & 88.4 & 83.9 & 98.0 & 82.6 & 88.2 \\
+ Class Bal.    & \ding{51} & \ding{51} & \ding{51} & \ding{51} & \ding{55} & 89.2 & 85.2 & 98.1 & 85.8 & 89.6 \\
+ Coherence     & \ding{51} & \ding{51} & \ding{51} & \ding{51} & \ding{51} & \textbf{89.7} & \textbf{86.1} & \textbf{98.2} & \textbf{86.6} & \textbf{90.2} \\
\bottomrule
\end{tabular}
\label{tab:1}
\end{table}

\begin{figure*}[h!]
    \centering

    \begin{minipage}[b]{0.45\textwidth}
        \centering
        \includegraphics[width=7.75cm, height=3.6cm]{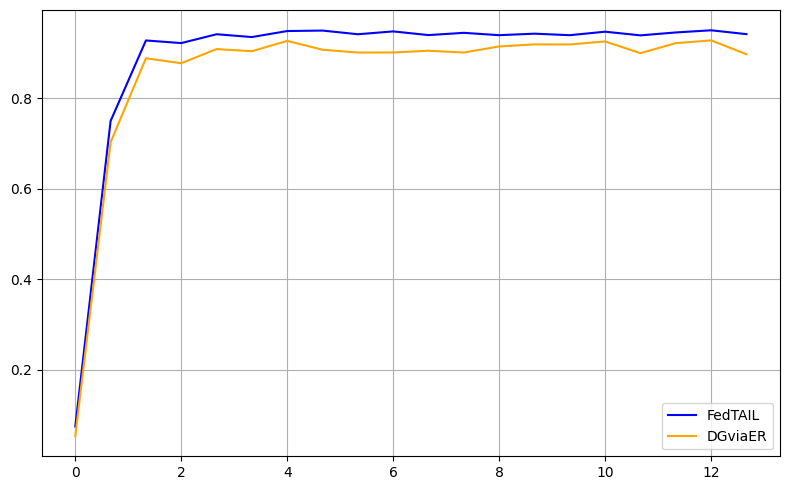}\\
        \textbf{Art}
    \end{minipage}
    \hfill
    \begin{minipage}[b]{0.45\textwidth}
        \centering
        \includegraphics[width=7.75cm, height=3.6cm]{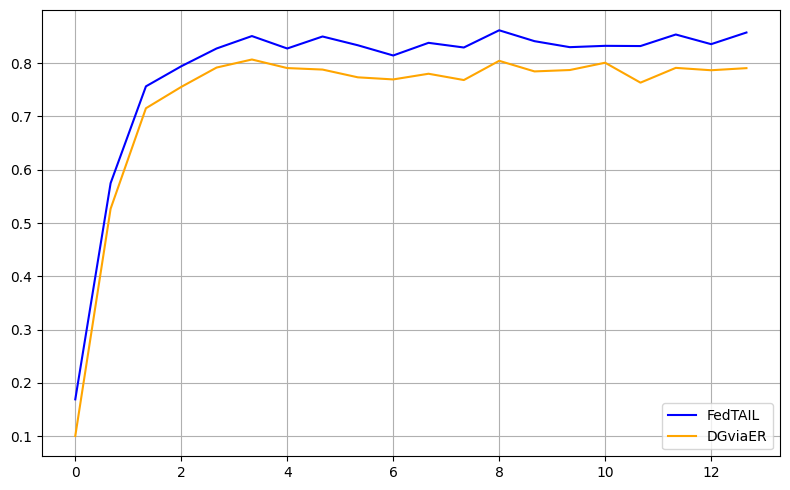}\\
        \textbf{Cartoon}
    \end{minipage}

    \vspace{0.15cm} 

    \begin{minipage}[b]{0.45\textwidth}
        \centering
        \includegraphics[width=7.75cm, height=3.6cm]{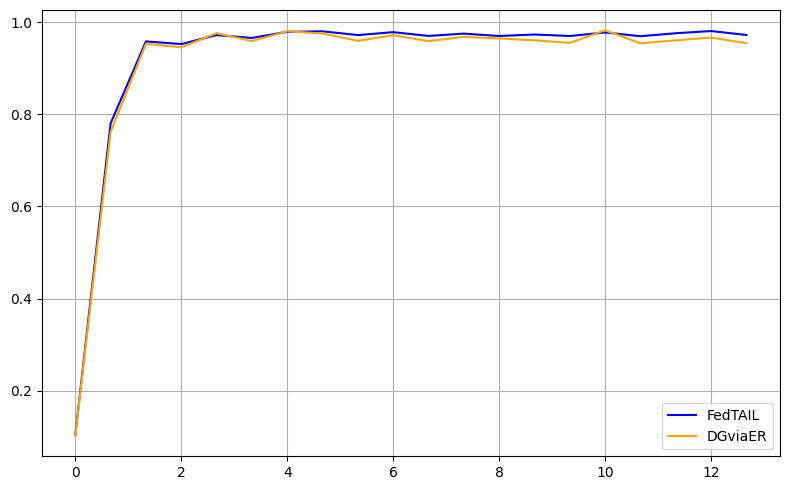}\\
        \textbf{Photo}
    \end{minipage}
    \hfill
    \begin{minipage}[b]{0.45\textwidth}
        \centering
        \includegraphics[width=7.75cm, height=3.6cm]{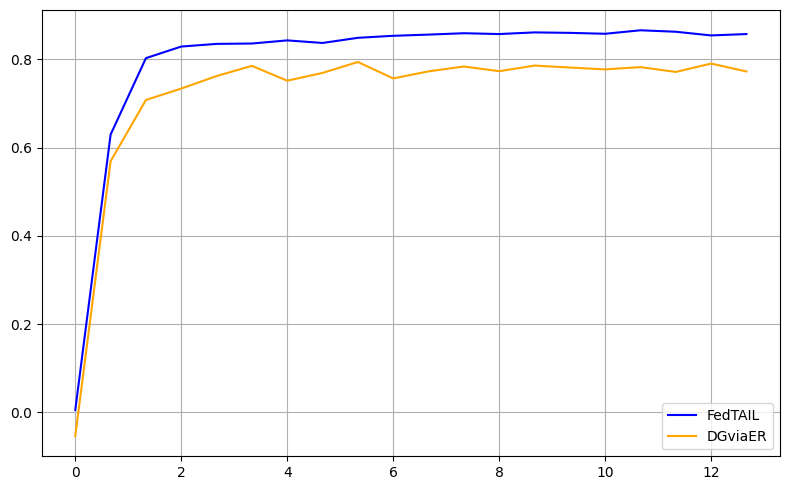}\\
        \textbf{Sketch}
    \end{minipage}

    \caption{
        Accuracy vs. Epoch comparison between \textbf{FedTAIL} and \textbf{DGviaER} across different domains (\textit{Art}, \textit{Cartoon}, \textit{Photo}, \textit{Sketch}) on the \textbf{PACS} dataset using ResNet-50.
    }
    \label{fig:1}
\end{figure*}

Figure~\ref{fig:1} illustrates the accuracy trends over training epochs for FedTAIL compared to the DGviaER baseline across different domains (\textit{Art}, \textit{Cartoon}, \textit{Photo}, \textit{Sketch}) on the PACS dataset. As shown, FedTAIL consistently achieves higher accuracy throughout the training process, across all domain splits. These trends confirm that FedTAIL facilitates faster convergence and improved robustness to domain shift. The learning curves demonstrate that our sharpness-guided, gradient-coherent optimization helps the model escape poor local minima early and promotes stable training dynamics, making it especially suited for federated scenarios where communication efficiency and convergence speed are critical.





\begin{figure*}[h!]
    \centering

    \includegraphics[width=0.245\textwidth]{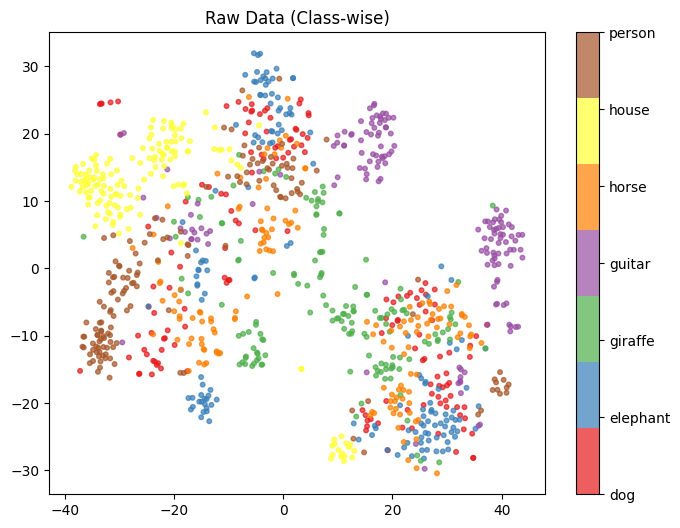}
    \includegraphics[width=0.245\textwidth]{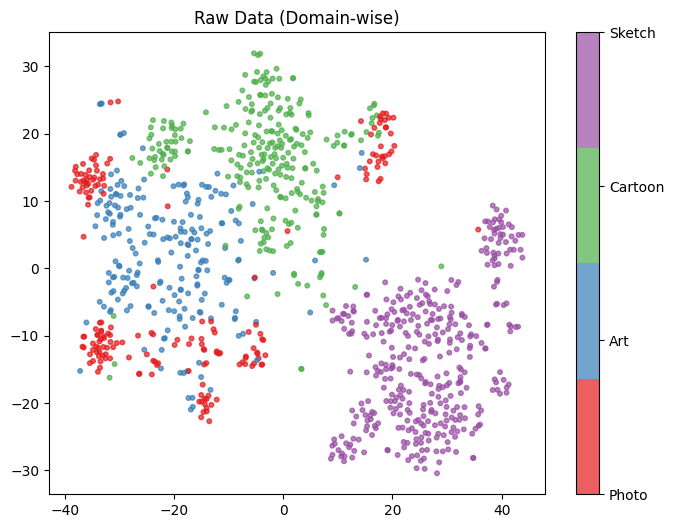}
    \includegraphics[width=0.245\textwidth]{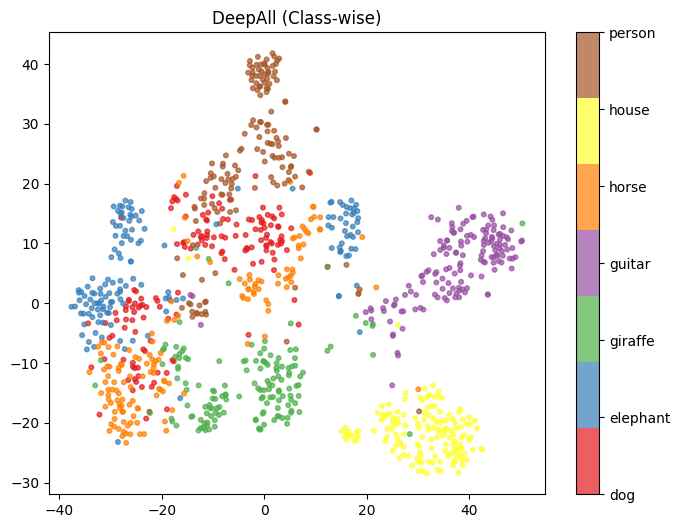}
    \includegraphics[width=0.245\textwidth]{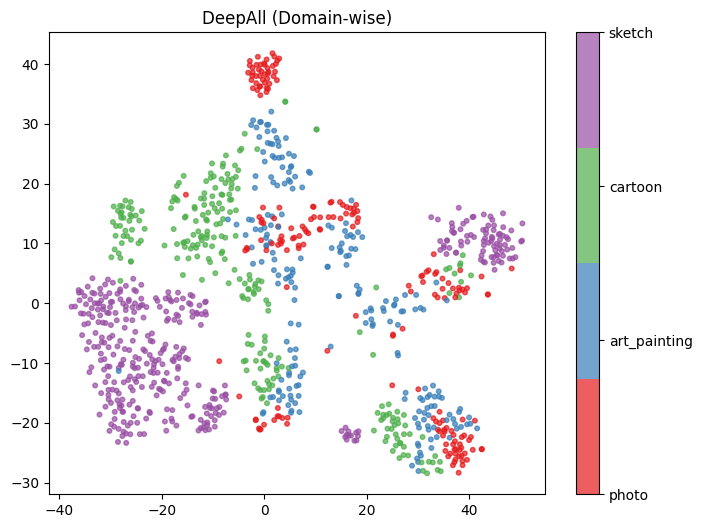}

    \vspace{0.3cm} 

    \includegraphics[width=0.245\textwidth]{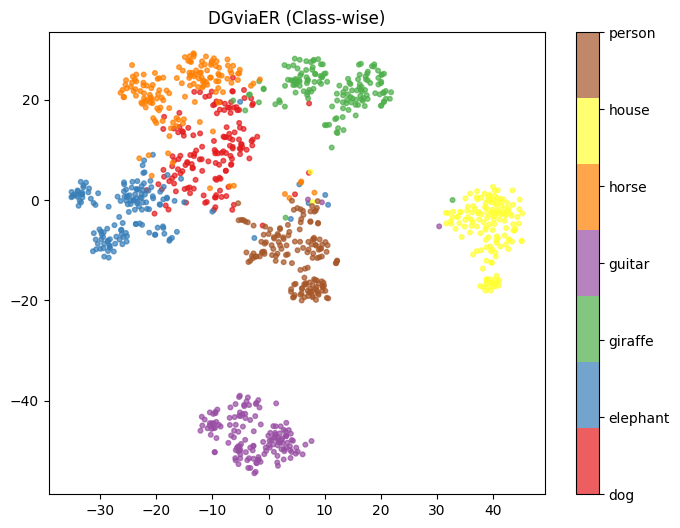}
    \includegraphics[width=0.245\textwidth]{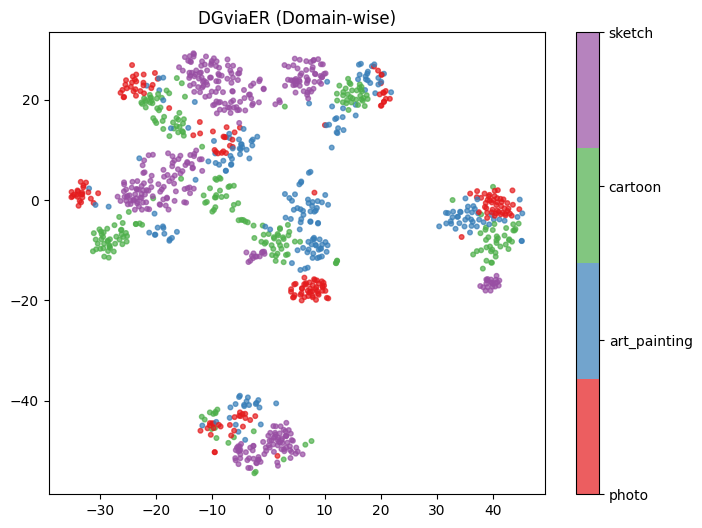}
    \includegraphics[width=0.245\textwidth]{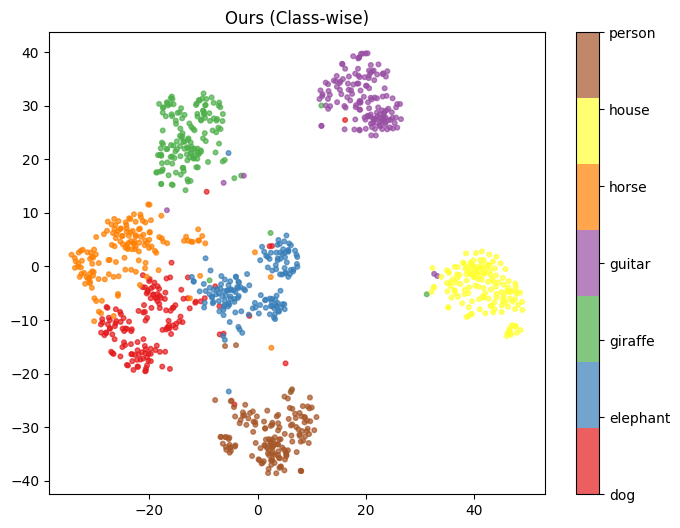}
    \includegraphics[width=0.245\textwidth]{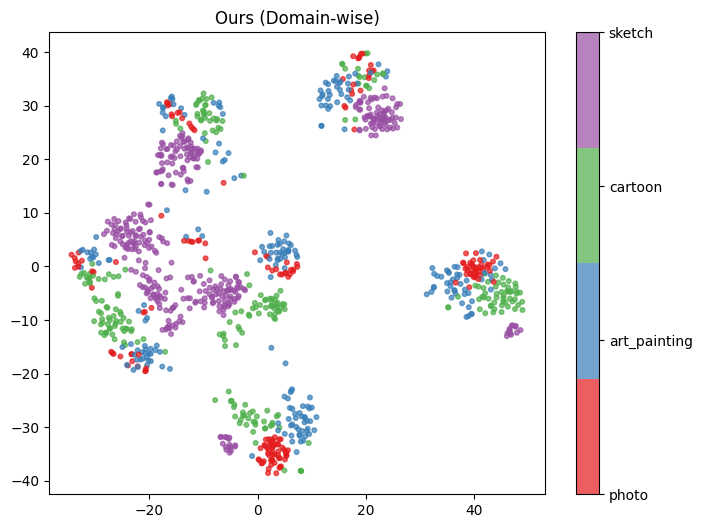}

    \caption{
        t-SNE visualizations of feature embeddings across class and domain. Top row: \textbf{Raw features} and \textbf{DeepAll}. Bottom row: \textbf{Standard DG} and the proposed \textbf{FedTAIL}. Left: class-wise separation. Right: domain-wise alignment. FedTAIL achieves better inter-class separability and cross-domain consistency.
    }
    \label{fig:2x4_grid}
\end{figure*}

To qualitatively assess representation quality, Figure~\ref{fig:2x4_grid} presents t-SNE visualizations of learned features across different methods. FedTAIL exhibits well-separated class clusters and improved domain alignment, compared to Raw features and DeepAll. The compact and consistent feature distributions reflect both strong intra-class cohesion and inter-domain alignment. This supports the role of class-wise sharpness minimization and entropy-regularized perturbations in achieving domain-agnostic and discriminative representations.

As shown in Figure~\ref{fig:2x4_grid}, FedTAIL consistently achieves higher accuracy than DGviaER across training epochs in both class-wise and domain-wise evaluations. These trends confirm that FedTAIL facilitates faster convergence and better robustness to domain shift over the course of training.

To provide further clarity on the sharpness-aware entropy regularization component ($\mathcal{L}_{\text{sharp-er}}$), we examine how the target-like predictive distribution $Q_T$ is constructed. Specifically, we estimate $Q_T$ based on class representation within each domain, calculated as the relative frequency of each class (i.e., \texttt{freq\_class / freq\_total}). For PACS, which exhibits clear domain-specific class imbalance, these distributions vary considerably across domains, as shown in Table~\ref{tab:2}. These $Q_T$ values are derived from a momentum-updated ensemble model and reflect domain-dependent prediction tendencies.

\begin{table}[h!]
\caption{\( Q_T \) values for PACS dataset across domains. Classes are ordered as: Dog, Elephant, Giraffe, Guitar, Horse, House, Person.}
\vspace{5pt}
\label{tab:2}
\centering
\begin{tabular}{lccccccc}
\toprule
\textbf{Domain} & Dog & Elephant & Giraffe & Guitar & Horse & House & Person \\
\midrule
Art Painting & 0.1851 & 0.1245 & 0.1392 & 0.0898 & 0.0981 & 0.1440 & 0.2192 \\
Cartoon      & 0.1660 & 0.1950 & 0.1476 & 0.0576 & 0.1382 & 0.1229 & 0.1728 \\
Photo        & 0.1132 & 0.1210 & 0.1090 & 0.1114 & 0.1192 & 0.1677 & 0.2587 \\
Sketch       & 0.1965 & 0.1883 & 0.1917 & 0.1547 & 0.2077 & 0.0204 & 0.0407 \\
\bottomrule
\end{tabular}
\end{table}

In contrast, Table~\ref{tab:3} reports $Q_T$ values for the Digits-DG dataset, which is inherently balanced across digit classes (0--9), resulting in a uniform $Q_T = 0.1$ across all domains. These insights validate that our formulation adapts well to both imbalanced and balanced data regimes.

\begin{table}[h!]
\caption{\( Q_T \) values for Digits-DG dataset across domains. All classes (0–9) have uniform \( Q_T = 0.1 \).}
\vspace{5pt}
\label{tab:3}
\centering
\begin{tabular}{lcccccccccc}
\toprule
\textbf{Domain} & 0 & 1 & 2 & 3 & 4 & 5 & 6 & 7 & 8 & 9 \\
\midrule
MNIST    & 0.1 & 0.1 & 0.1 & 0.1 & 0.1 & 0.1 & 0.1 & 0.1 & 0.1 & 0.1 \\
MNIST-M  & 0.1 & 0.1 & 0.1 & 0.1 & 0.1 & 0.1 & 0.1 & 0.1 & 0.1 & 0.1 \\
SVHN     & 0.1 & 0.1 & 0.1 & 0.1 & 0.1 & 0.1 & 0.1 & 0.1 & 0.1 & 0.1 \\
SYN      & 0.1 & 0.1 & 0.1 & 0.1 & 0.1 & 0.1 & 0.1 & 0.1 & 0.1 & 0.1 \\
\bottomrule
\end{tabular}
\end{table}

Importantly, we applied the same frequency-based $Q_T$ estimation approach to Office-Home and DomainNet datasets. Due to the large number of classes in these datasets, their tables are omitted for brevity. Nonetheless, these results collectively demonstrate that our sharpness-aware entropy regularization mechanism generalizes effectively across datasets with diverse class distributions and domain characteristics. This capability further confirms FedTAIL’s robustness to real-world federated learning challenges, where heterogeneity in class presence and data distributions is the norm rather than the exception.

\bibliographystyle{unsrt}  
\bibliography{fedtail}

\begin{thebibliography}{10}

\bibitem{zhang2022towards}
Hanlin Zhang, Yi-Fan Zhang, Weiyang Liu, Adrian Weller, Bernhard Sch{\"o}lkopf, and Eric~P Xing.
\newblock Towards principled disentanglement for domain generalization.
\newblock In {\em IEEE/CVF Conference on Computer Vision and Pattern Recognition (CVPR)}, pages 8014--8024, 2022.

\bibitem{qiao2020learning}
Fengchun Qiao, Long Zhao, and Xi~Peng.
\newblock Learning to learn single domain generalization.
\newblock In {\em Proceedings of the IEEE/CVF conference on computer vision and pattern recognition}, pages 12556--12565, 2020.

\bibitem{balaji2018metareg}
Y.~Balaji, S.~Sankaranarayanan, and R.~Chellappa.
\newblock Metareg: Towards domain generalization using meta-regularization.
\newblock In {\em NeurIPS}, 2018.

\bibitem{muandet2013domain}
Krikamol Muandet, David Balduzzi, and Bernhard Scholkopf.
\newblock Domain generalization via invariant feature representation.
\newblock In {\em International Conference on Machine Learning}, pages 10--18. PMLR, 2013.

\bibitem{li2018learning}
Da~Li, Yongxin Yang, Yi-Zhe Song, and Timothy Hospedales.
\newblock Learning to generalize: Meta-learning for domain generalization.
\newblock In {\em Proceedings of the AAAI conference on artificial intelligence}, volume~32, 2018.

\bibitem{wang2022semantic}
Mengzhu Wang, Jianlong Yuan, Qi~Qian, Zhibin Wang, and Hao Li.
\newblock Semantic data augmentation based distance metric learning for domain generalization.
\newblock In {\em Proceedings of the 30th ACM international conference on multimedia}, pages 3214--3223, 2022.

\bibitem{zhao2020domain}
Shanshan Zhao, Mingming Gong, Tongliang Liu, Huan Fu, and Dacheng Tao.
\newblock Domain generalization via entropy regularization.
\newblock {\em Advances in neural information processing systems}, 33:16096--16107, 2020.

\bibitem{foret2020sharpness}
Pierre Foret, Ariel Kleiner, Hossein Mobahi, and Behnam Neyshabur.
\newblock Sharpness-aware minimization for efficiently improving generalization.
\newblock {\em arXiv preprint arXiv:2010.01412}, 2020.

\bibitem{rangwani2022escaping}
Harsh Rangwani, Sumukh~K Aithal, Mayank Mishra, et~al.
\newblock Escaping saddle points for effective generalization on class-imbalanced data.
\newblock {\em Advances in Neural Information Processing Systems}, 35:22791--22805, 2022.

\bibitem{chen2019domain}
Minghao Chen, Hongyang Xue, and Deng Cai.
\newblock Domain adaptation for semantic segmentation with maximum squares loss.
\newblock In {\em Proceedings of the IEEE/CVF international conference on computer vision}, pages 2090--2099, 2019.

\bibitem{grandvalet2004semi}
Yves Grandvalet and Yoshua Bengio.
\newblock Semi-supervised learning by entropy minimization.
\newblock {\em Advances in neural information processing systems}, 17, 2004.

\bibitem{ganin2016domain}
Yaroslav Ganin, Evgeniya Ustinova, Hana Ajakan, Pascal Germain, Hugo Larochelle, Fran{\c{c}}ois Laviolette, Mario March, and Victor Lempitsky.
\newblock Domain-adversarial training of neural networks.
\newblock {\em Journal of machine learning research}, 17(59):1--35, 2016.

\bibitem{li2018domain}
Haoliang Li, Sinno~Jialin Pan, Shiqi Wang, and Alex~C Kot.
\newblock Domain generalization with adversarial feature learning.
\newblock In {\em CVPR}, pages 5400--5409, 2018.

\bibitem{bahng2020learning}
Hyojin Bahng, Sanghyuk Chun, Sangdoo Yun, Jaegul Choo, and Seong~Joon Oh.
\newblock Learning de-biased representations with biased representations.
\newblock In {\em International conference on machine learning}, pages 528--539. PMLR, 2020.

\bibitem{zhang2021adaptive}
Marvin Zhang, Henrik Marklund, Nikita Dhawan, Abhishek Gupta, Sergey Levine, and Chelsea Finn.
\newblock Adaptive risk minimization: Learning to adapt to domain shift.
\newblock {\em Advances in Neural Information Processing Systems}, 34:23664--23678, 2021.

\bibitem{dou2019domain}
Qi~Dou, Daniel Coelho~de Castro, Konstantinos Kamnitsas, and Ben Glocker.
\newblock Domain generalization via model-agnostic learning of semantic features.
\newblock In {\em NeurIPS}, volume~32, 2019.

\bibitem{zhou2021domain}
Kaiyang Zhou, Yongxin Yang, Yu~Qiao, and Tao Xiang.
\newblock Domain adaptive ensemble learning.
\newblock {\em IEEE Transactions on Image Processing}, 30:8008--8018, 2021.

\bibitem{shankar2018generalizing}
Shiv Shankar, Vihari Piratla, Soumen Chakrabarti, Siddhartha Chaudhuri, Preethi Jyothi, and Sunita Sarawagi.
\newblock Generalizing across domains via cross-gradient training.
\newblock In {\em International Conference on Learning Representations (ICLR)}, 2018.

\bibitem{carlucci2019domain}
Fabio~M Carlucci, Antonio D’Innocente, Silvia Bucci, Barbara Caputo, and Tatiana Tommasi.
\newblock Domain generalization by solving jigsaw puzzles.
\newblock In {\em CVPR}, pages 2229--2238, 2019.

\bibitem{peng2019domain}
Xingchao Peng, Zijun Huang, Ximeng Sun, and Kate Saenko.
\newblock Domain agnostic learning with disentangled representations.
\newblock In {\em ICML}, pages 5102--5112, 2019.

\bibitem{khosla2012undoing}
Aditya Khosla, Tinghui Zhou, Tomasz Malisiewicz, Alexei~A Efros, and Antonio Torralba.
\newblock Undoing the damage of dataset bias.
\newblock In {\em Computer Vision--ECCV 2012: 12th European Conference on Computer Vision, Florence, Italy, October 7-13, 2012, Proceedings, Part I 12}, pages 158--171. Springer, 2012.

\bibitem{wang2020cross}
Guoqing Wang, Hu~Han, Shiguang Shan, and Xilin Chen.
\newblock Cross-domain face presentation attack detection via multi-domain disentangled representation learning.
\newblock In {\em Proceedings of the IEEE/CVF conference on computer vision and pattern recognition}, pages 6678--6687, 2020.

\bibitem{krueger2021out}
David Krueger, Ethan Caballero, Joern-Henrik Jacobsen, Amy Zhang, Jonathan Binas, Dinghuai Zhang, Remi Le~Priol, and Aaron Courville.
\newblock Out-of-distribution generalization via risk extrapolation (rex).
\newblock In {\em International conference on machine learning}, pages 5815--5826. PMLR, 2021.

\bibitem{arjovsky2019invariant}
Martin Arjovsky, L{\'e}on Bottou, Ishaan Gulrajani, and David Lopez-Paz.
\newblock Invariant risk minimization.
\newblock {\em arXiv preprint arXiv:1907.02893}, 2019.

\bibitem{mansilla2021domain}
Lucas Mansilla, Rodrigo Echeveste, Diego~H Milone, and Enzo Ferrante.
\newblock Domain generalization via gradient surgery.
\newblock In {\em Proceedings of the IEEE/CVF international conference on computer vision}, pages 6630--6638, 2021.

\bibitem{liu2022towards}
Yong Liu, Siqi Mai, Xiangning Chen, Cho-Jui Hsieh, and Yang You.
\newblock Towards efficient and scalable sharpness-aware minimization.
\newblock In {\em Proceedings of the IEEE/CVF Conference on Computer Vision and Pattern Recognition}, pages 12360--12370, 2022.

\bibitem{du2021efficient}
Jiawei Du, Hanshu Yan, Jiashi Feng, Joey~Tianyi Zhou, Liangli Zhen, Rick Siow~Mong Goh, and Vincent~YF Tan.
\newblock Efficient sharpness-aware minimization for improved training of neural networks.
\newblock {\em arXiv preprint arXiv:2110.03141}, 2021.

\bibitem{hochreiter1994simplifying}
Sepp Hochreiter and J{\"u}rgen Schmidhuber.
\newblock Simplifying neural nets by discovering flat minima.
\newblock {\em Advances in neural information processing systems}, 7, 1994.

\bibitem{keskar2016large}
Nitish~Shirish Keskar, Dheevatsa Mudigere, Jorge Nocedal, Mikhail Smelyanskiy, and Ping Tak~Peter Tang.
\newblock On large-batch training for deep learning: Generalization gap and sharp minima.
\newblock {\em arXiv preprint arXiv:1609.04836}, 2016.

\bibitem{dinh2017sharp}
Laurent Dinh, Razvan Pascanu, Samy Bengio, and Yoshua Bengio.
\newblock Sharp minima can generalize for deep nets.
\newblock In {\em International Conference on Machine Learning}, pages 1019--1028. PMLR, 2017.

\bibitem{cha2021swad}
Junbum Cha, Sanghyuk Chun, Kyungjae Lee, Han-Cheol Cho, Seunghyun Park, Yunsung Lee, and Sungrae Park.
\newblock Swad: Domain generalization by seeking flat minima.
\newblock {\em Advances in Neural Information Processing Systems}, 34:22405--22418, 2021.

\bibitem{wang2020learning}
Shujun Wang, Lequan Yu, Caizi Li, Chi-Wing Fu, and Pheng-Ann Heng.
\newblock Learning from extrinsic and intrinsic supervisions for domain generalization.
\newblock In {\em European Conference on Computer Vision}, pages 159--176. Springer, 2020.

\bibitem{su2024sharpness}
Houcheng Su, Weihao Luo, Daixian Liu, Mengzhu Wang, Jing Tang, Junyang Chen, Cong Wang, and Zhenghan Chen.
\newblock Sharpness-aware model-agnostic long-tailed domain generalization.
\newblock In {\em Proceedings of the AAAI Conference on Artificial Intelligence}, volume~38, pages 15091--15099, 2024.

\bibitem{ganin2015unsupervised}
Yaroslav Ganin and Victor Lempitsky.
\newblock Unsupervised domain adaptation by backpropagation.
\newblock In {\em ICML}, pages 1180--1189, 2015.

\bibitem{wang2023sharpness}
Pengfei Wang, Zhaoxiang Zhang, Zhen Lei, and Lei Zhang.
\newblock Sharpness-aware gradient matching for domain generalization.
\newblock In {\em Proceedings of the IEEE/CVF Conference on Computer Vision and Pattern Recognition (CVPR)}, 2023.

\bibitem{li2017deeper}
D.~Li, Y.~Yang, Y.-Z. Song, and T.~M. Hospedales.
\newblock Deeper, broader and artier domain generalization.
\newblock In {\em ICCV}, 2017.

\bibitem{venkateswara2017deep}
Hemanth Venkateswara, Jose Eusebio, Shayok Chakraborty, and Sethuraman Panchanathan.
\newblock Deep hashing network for unsupervised domain adaptation.
\newblock In {\em IEEE Conference on Computer Vision and Pattern Recognition (CVPR)}, pages 5018--5027, 2017.

\bibitem{zhou2020learning}
Kaiyang Zhou, Yongxin Yang, Timothy~M Hospedales, and Tao Xiang.
\newblock Learning to generate novel domains for domain generalization.
\newblock In {\em European Conference on Computer Vision (ECCV)}, pages 561--578, 2020.

\bibitem{peng2019moment}
X.~Peng, Q.~Bai, X.~Xia, Z.~Huang, K.~Saenko, and B.~Wang.
\newblock Moment matching for multi-source domain adaptation.
\newblock In {\em ICCV}, 2019.

\bibitem{dinnocente2018domain}
A.~D’Innocente and B.~Caputo.
\newblock Domain generalization with domain-specific aggregation modules.
\newblock In {\em GCPR}, 2018.

\bibitem{vapnik1998statistical}
Vladimir~Naumovich Vapnik.
\newblock {\em Statistical learning theory}.
\newblock Wiley-Interscience, 1998.

\bibitem{li2022epi}
Yuhang Li, Yaqing Wang, Yifan Li, Yuxin Li, Yujie Zhang, and Liang Wang.
\newblock Epi-fcr: Episodic fine-grained cross-domain few-shot learning via feature calibration and relation alignment.
\newblock In {\em Proceedings of the 30th ACM International Conference on Multimedia}, pages 1234--1243. ACM, 2022.

\bibitem{wang2020heterogeneous}
Yufei Wang, Haoliang Li, and Alex~C. Kot.
\newblock Heterogeneous domain generalization via domain mixup.
\newblock {\em arXiv preprint arXiv:2009.05448}, 2020.

\bibitem{deepall}
Da~Li, Yongxin Yang, Yi-Zhe Song, and Timothy~M. Hospedales.
\newblock Deeper, broader and artier domain generalization.
\newblock In {\em Proceedings of the IEEE International Conference on Computer Vision (ICCV)}, pages 5542--5550, Oct 2017.

\bibitem{wang2021embracing}
Yufei Wang, Haoliang Li, Lap-pui Chau, and Alex~C Kot.
\newblock Embracing the dark knowledge: Domain generalization using regularized knowledge distillation.
\newblock In {\em ACM International Conference on Multimedia}, pages 2595--2604, 2021.

\bibitem{zhao2024symmetric}
Di~Zhao, Yun~Sing Koh, Gillian Dobbie, Hongsheng Hu, and Philippe Fournier-Viger.
\newblock Symmetric self-paced learning for domain generalization.
\newblock In {\em AAAI Conference on Artificial Intelligence}, volume~38, pages 16961--16969, 2024.

\bibitem{dger}
Shiqi Zhao, Mingkui Gong, Ting Liu, Yunchao Fu, and Dacheng Tao.
\newblock Domain generalization via entropy regularization.
\newblock In {\em Advances in Neural Information Processing Systems (NeurIPS)}, volume~33, pages 16096--16107, 2020.

\bibitem{zhou2020deep}
Kaiyang Zhou, Yongxin Yang, Timothy~M Hospedales, and Tao Xiang.
\newblock Deep domain-adversarial image generation for domain generalisation.
\newblock In {\em AAAI Conference on Artificial Intelligence (AAAI)}, pages 13025--13032, 2020.

\bibitem{huang2023sentence}
Zeyi Huang, Andy Zhou, Zijian Ling, Mu~Cai, Haohan Wang, and Yong~Jae Lee.
\newblock A sentence speaks a thousand images: Domain generalization through distilling clip with language guidance.
\newblock In {\em International Conference on Computer Vision}, pages 11685--11695, 2023.

\bibitem{huang2020self}
Zeyi Huang, Haohan Wang, Eric~P Xing, and Dong Huang.
\newblock Self-challenging improves cross-domain generalization.
\newblock In {\em ECCV}, pages 124--140. Springer, 2020.

\bibitem{xu2021fourier}
Qinwei Xu, Ruipeng Zhang, Ya~Zhang, Yanfeng Wang, and Qi~Tian.
\newblock A fourier-based framework for domain generalization.
\newblock In {\em IEEE/CVF Conference on Computer Vision and Pattern Recognition (CVPR)}, pages 14383--14392, 2021.

\bibitem{mahajan2021domain}
Divyat Mahajan, Shruti Tople, and Amit Sharma.
\newblock Domain generalization using causal matching.
\newblock In {\em ICML}, pages 7313--7324, 2021.

\bibitem{jeon2021feature}
Seogkyu Jeon, Kibeom Hong, Pilhyeon Lee, Jewook Lee, and Hyeran Byun.
\newblock Feature stylization and domain-aware contrastive learning for domain generalization.
\newblock In {\em ACM Multimedia}, pages 22--31, 2021.

\bibitem{ffdi}
Jingye Wang, Ruoyi Du, Dongliang Chang, Kongming Liang, and Zhanyu Ma.
\newblock Domain generalization via frequency-domain-based feature disentanglement and interaction.
\newblock In {\em Proceedings of the 30th ACM International Conference on Multimedia}, MM '22, page 4821–4829, New York, NY, USA, 2022. Association for Computing Machinery.

\bibitem{yao2022pcl}
Xufeng Yao, Yang Bai, Xinyun Zhang, Yuechen Zhang, Qi~Sun, Ran Chen, Ruiyu Li, and Bei Yu.
\newblock Pcl: Proxy-based contrastive learning for domain generalization.
\newblock In {\em IEEE/CVF Conference on Computer Vision and Pattern Recognition (CVPR)}, pages 7087--7097, 2022.

\bibitem{kang2022style}
Juwon Kang, Sohyun Lee, Namyup Kim, and Suha Kwak.
\newblock Style neophile: Constantly seeking novel styles for domain generalization.
\newblock In {\em Proceedings of the IEEE/CVF Conference on Computer Vision and Pattern Recognition (CVPR)}, pages 7130--7140, June 2022.

\bibitem{lv2023improving}
Fangrui Lv, Jian Liang, Shuang Li, Jinming Zhang, and Di~Liu.
\newblock Improving generalization with domain convex game.
\newblock In {\em Proceedings of the IEEE/CVF Conference on Computer Vision and Pattern Recognition}, pages 24315--24324, 2023.

\bibitem{he2016resnet}
Kaiming He, Xiangyu Zhang, Shaoqing Ren, and Jian Sun.
\newblock Deep residual learning for image recognition.
\newblock {\em CoRR}, 2015.

\bibitem{deng2009imagenet}
J.~Deng, W.~Dong, R.~Socher, L.J. Li, K.~Li, and L.~Fei-Fei.
\newblock Imagenet: A large-scale hierarchical image database.
\newblock In {\em CVPR}, 2009.

\bibitem{rise}
Zeyi Huang, Andy Zhou, Zijian Lin, Mu~Cai, Haohan Wang, and Yong~Jae Lee.
\newblock A sentence speaks a thousand images: Domain generalization through distilling clip with language guidance.
\newblock {\em arXiv preprint arXiv:2309.12530}, 2023.

\end{thebibliography}

\end{document}